\documentclass[letterpaper]{article} % DO NOT CHANGE THIS

\usepackage{aaai2027}
\usepackage[hyphens]{url}  % DO NOT CHANGE THIS
\usepackage{graphicx} % DO NOT CHANGE THIS
\usepackage{natbib}  % DO NOT CHANGE THIS AND DO NOT ADD ANY OPTIONS TO IT
\usepackage{caption} % DO NOT CHANGE THIS AND DO NOT ADD ANY OPTIONS TO IT
\usepackage{amssymb}
\usepackage{algorithm}
\usepackage{algorithmic}
\usepackage{amsmath}
\usepackage{newfloat}
\usepackage{listings}
\DeclareCaptionStyle{ruled}{labelfont=normalfont,labelsep=colon,strut=off} % DO NOT CHANGE THIS
\floatstyle{ruled}
\newfloat{listing}{tb}{lst}{}
\floatname{listing}{Listing}

\usepackage{booktabs}

\title{PluginEval: A Diagnostic Benchmark for \\Fine-Grained Error Attribution in Function Calling}

\author{
    Dongjie Xu\equalcontrib\textsuperscript{\rm 1},
    Julius\equalcontrib\textsuperscript{\rm 2},
    Hanchi Dong\equalcontrib\textsuperscript{\rm 3},
    Minghua Tang\textsuperscript{\rm 4},
    Yuxuan Sun\textsuperscript{\rm 3},
    Ziwei Nie\textsuperscript{\rm 5},\\
    Zicheng Liu\textsuperscript{\rm 6},
    Dujun Qing\textsuperscript{\rm 7},
    Jiajie Xu\textsuperscript{\rm 1}
}
\affiliations{
    \textsuperscript{\rm 1}Soochow University\\
    \textsuperscript{\rm 2}Tencent Inc.\\
    \textsuperscript{\rm 3}Beijing University of Posts and Telecommunications\\
    \textsuperscript{\rm 4}Jiangnan University\\
    \textsuperscript{\rm 5}Nanjing University\\
    \textsuperscript{\rm 6}Xiaohongshu Inc.\\
    \textsuperscript{\rm 7}University of Chinese Academy of Sciences
}

\begin{document}

\maketitle

\begin{abstract}
% TODO: Final abstract (~150-200 words). Draft skeleton below.
% Reliable evaluation of function calling is essential as large language models increasingly rely on external tools to execute complex tasks and operate as agents. However, current benchmarks provide an incomplete account of this capability: aggregate scores conceal distinct failure modes, test queries often fail to distinguish strong models, and scoring protocols are rarely validated against human judgments.

Reliable evaluation of tool routing is critical as Large Language Models increasingly operate as autonomous agents. Current benchmarks face three structural limitations: data distributions that follow a power law leave rare scenarios underrepresented; the absence of adversarial hard negatives obscures performance differences across models; and annotation pipelines depend on LLM judgments that have not been validated through execution. In this paper, we introduce PluginEval, a benchmark constructed through a two-stage framework that systematically mitigates these limitations. First, we formulate tool routing as a sequence of three decisions and separate generation from verification. LLMs propose candidate calls, while deterministic validation and real API execution provide reliable quality signals. Second, we decompose each plugin by capability, intent, and boundary to identify trigger and exclusion scenarios. We then generate queries at different difficulty levels to fill coverage gaps, including adversarial negatives targeting three failure modes, and return them to the first stage for annotation. This process creates a closed loop that iterates until coverage converges. For evaluation, we move beyond aggregate accuracy. An LLM judge anchored to gold annotations classifies failures as missed calls, spurious calls, or parameter errors, producing a detailed error profile for each model. We evaluate five model families, including proprietary models and models with open weights, analyze their performance across difficulty levels and error categories, and validate the judge through agreement with human annotations.

\end{abstract}

% Uncomment the following to link to your code, datasets, an extended version or similar.
% You must keep this block between (not within) the abstract and the main body of the paper.
% Make sure that you do not de-anonymize yourself with these links.
% \begin{links}
%     \link{Code}{https://aaai.org/example/code}
%     \link{Datasets}{https://aaai.org/example/datasets}
%     \link{Extended version}{https://aaai.org/example/extended-version}
% \end{links}

\begin{figure}[t] \centering \includegraphics[width=0.75\columnwidth]{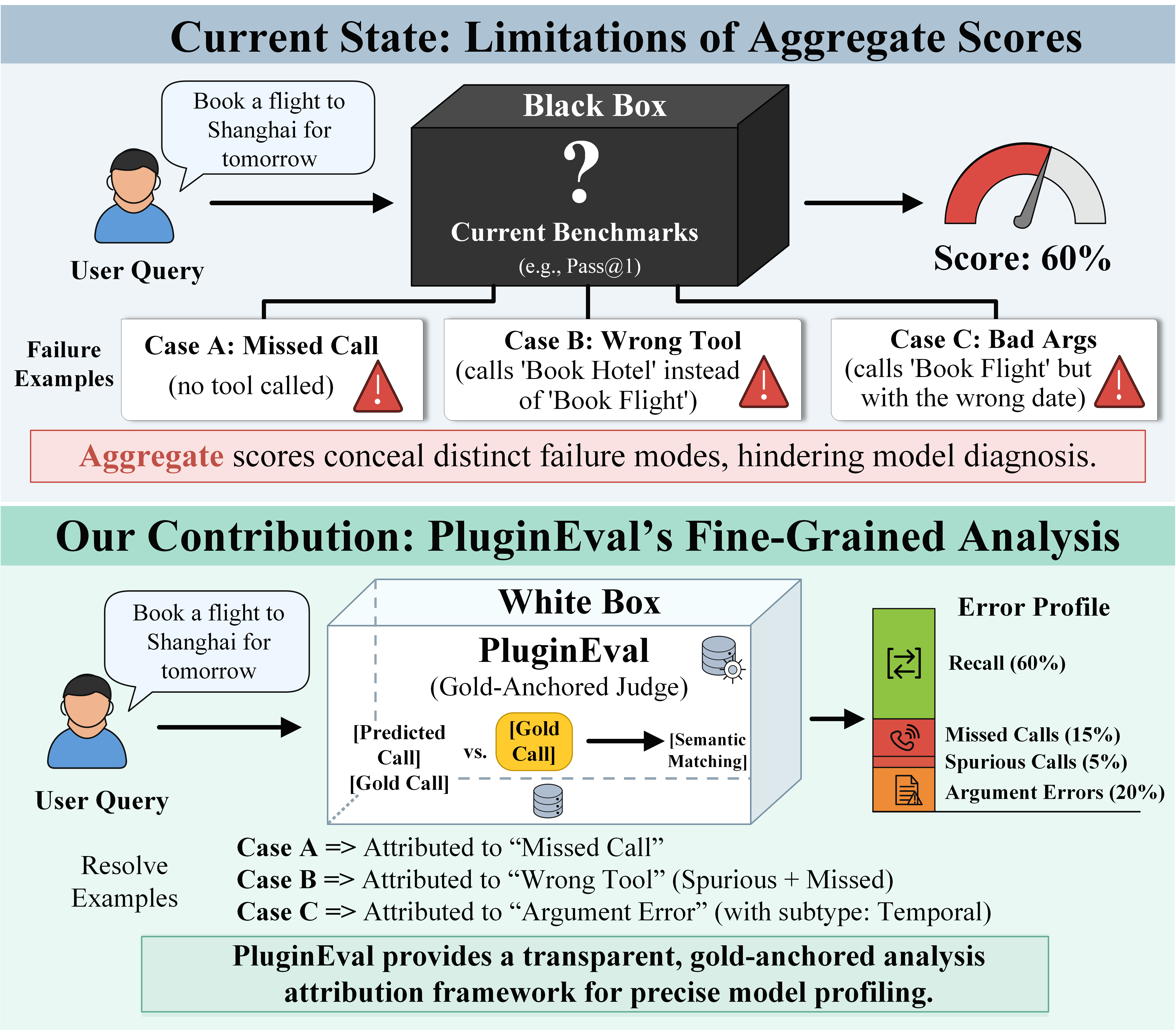} \caption{Motivation for fine grained function calling evaluation: aggregate scores obscure distinct failure modes.} \label{fig:motivation} \end{figure}

\section{Introduction}
% 第一段任务介绍和背景（8行左右） 第二段是现有工作分析（总结相关工作部分的不足15行）
% 第三段引出challenge（2-3点） 第四段针对challenge得方法设计 最后contribution（两点针对评测样本生产 一点是实验）
\label{sec:introduction}
Large language models increasingly use external tools through function calls \citep{yao2022react}. Reliable tool routing requires three hierarchical decisions: whether a tool is needed, whether the target plugin is relevant, and whether the request can be grounded into a valid invocation or should be rejected. Errors at an earlier level invalidate correct downstream decisions. Tool routing benchmarks should therefore evaluate tool necessity, plugin relevance, invocation feasibility, and argument grounding rather than call syntax alone.

% Existing work has broadened tool use evaluation to function relevance, abstention, and agentic interaction, while automated pipelines improve data quality through structural checks, model based verification, and actual execution \citep{li2023api,qin2024toolllm,patil2025berkeley,liu2024apigen,liu2025toolace}. Yet structural validity and execution success do not ensure semantic correctness, and LLM evaluators may introduce systematic biases \citep{liu2023g,zheng2023judging,wataoka2024self}. Dataset scale alone neither reveals nor closes structured coverage gaps across plugin specific trigger and exclusion boundaries. Moreover, construction heuristics may not reflect the difficulty observed from actual routing behavior. Addressing these limitations requires a benchmark construction process that connects reliable annotation, structured coverage analysis, and empirical difficulty calibration.

Benchmarks evaluate function relevance, abstention, and agentic interaction, while automated pipelines apply structural checks, model verification, and execution \citep{li2023api,qin2024toolllm,patil2025berkeley,liu2024apigen,liu2025toolace}. Yet most target English and poorly cover Chinese queries with ambiguous, implicit, or context dependent intent. Structural validity and execution success do not ensure semantic correctness, while LLM judges may be biased \citep{liu2023g,zheng2023judging,wataoka2024self}. Larger datasets still miss the trigger and exclusion boundaries of plugins, and heuristic difficulty labels may diverge from routing behavior. These gaps motivate a Chinese benchmark combining reliable annotation, structured coverage, and difficulty calibration based on model behavior.

This requirement gives rise to three technical challenges. First, trustworthy annotation must distinguish valid calls, justified abstentions, and plausible but incorrect proposals by reconciling semantic, structural, and execution evidence. Second, scenario level coverage must recover latent trigger and exclusion boundaries from plugin specifications and account for queries that match multiple scenarios without inflating coverage. Third, difficulty alignment must reconcile type specific construction criteria with router dependent behavior. These challenges follow a natural dependency: coverage can be measured only over verified examples, synthesis targets must be derived from the resulting coverage gaps, and difficulty can be calibrated only after synthesized examples are evaluated by routing models.

We introduce PluginEval and its Closed-Loop Construction Framework (PCCF), which follows this dependency structure. Stage I first establishes trustworthy annotations by selecting executable calls or justified abstentions through schema validation, independent LLM evaluation, and tool execution. Failed verification traces are reused to refine subsequent proposals. Utilizing only validated samples, Stage II derives plugin-level trigger scenarios, constructs a hierarchical coverage matrix, and synthesizes examples for undercovered units. It then calibrates difficulty of negative examples through repeated routing evaluation and uses recognition evidence to strengthen negatives that remain below the target difficulty. The two stages form a closed loop in which verification outcomes drive annotation repair, while coverage deficits and routing outcomes drive targeted synthesis. For evaluation, our gold-anchored judge aligns model predictions with verified gold calls and attributes failures to miss recall, parameter errors, and over recall.

Our contributions are summarized as follows:
\begin{itemize}
    %\item We formulate tool routing as a hierarchical decision process over tool necessity, target plugin relevance, and invocation feasibility with argument grounding. This formulation enables level specific diagnosis of routing failures and underlies PluginEval, which covers explicit trigger and exclusion scenarios, three routing boundary negative types, and behaviorally calibrated difficulty.

    \item We formulate tool routing as a three-level decision process over tool necessity, plugin relevance, and valid invocation with grounded arguments. Based on this formulation, we construct PluginEval with explicit trigger and exclusion scenarios, three types of boundary negatives, and behaviorally calibrated difficulty.

    %\item We propose PCCF, a closed loop benchmark construction framework that decouples annotation trustworthiness from scenario neededness. PCCF converts schema, evaluation, and execution outcomes into annotation repair signals, while using coverage deficits and observed routing behavior to drive targeted synthesis and difficulty refinement.
    
    \item We propose PCCF, a closed loop framework that separates annotation reliability from coverage requirements. PCCF uses schema validation, independent evaluation, and tool execution to repair annotations, while coverage gaps and observed routing behavior guide targeted synthesis and difficulty refinement.

    %\item We conduct a comprehensive evaluation of five frontier LLMs using our gold-anchored framework for fine-grained error attribution. The results reveal substantial performance variation across difficulty distributions and distinct failure profiles across models. Human audits further confirm the reliability of both the benchmark annotations and the attribution judge.

    \item We evaluate five frontier LLMs using a gold-anchored framework for fine-grained error attribution. The results reveal substantial sensitivity to difficulty composition and distinct failure profiles across models, while human audits confirm the reliability of both the benchmark annotations and the attribution judge.
\end{itemize}

% Our contributions are as follows:
% \begin{itemize}
%     \item A methodology for constructing high-discrimination hard samples (difficulty grading, discrimination measurement, hard negatives, and scenario-grounded long-tail positives). % TODO
%     \item A fine-grained error-attribution framework (missed / spurious calls and inaccurate arguments) driven by a gold-anchored LLM judge. % TODO
%     \item Evidence that the framework agrees closely with human judgment ($\kappa$ / $\rho$). % TODO
%     \item A cross-model evaluation of $N$ mainstream models with error-profile findings. % TODO
% \end{itemize}

% 1.5 A memorable number (e.g., on our hard negatives, SOTA model X reaches
% error type Y at Z%). TODO after experiments.

\section{Related Work}
\paragraph{Tool-use and function-calling benchmarks.}
Tool-use benchmarks evaluate API selection and invocation. Toolformer and Gorilla/APIBench study tool acquisition from demonstrations, documentation, or self-supervision \cite{schick2023toolformer,patil2024gorilla}, while ToolLLM/ToolBench scales to real-world RESTful APIs \cite{qin2024toolllm}. Executable benchmarks broaden coverage: BFCL covers single-/multi-turn, parallel, and abstention cases with AST matching \cite{patil2025berkeley}; FuncBenchGen controls dependency depth and distractors \cite{maekawa2025towards}; HammerBench targets imperfect mobile-assistant contexts \cite{wang2024hammerbench}; and NESTFUL tests nested API-call sequences \cite{basu2025nestful}. Recent multilingual and realistic-API benchmarks add MLCL/Lost in Execution for execution-interface vs. semantic errors \cite{luo2026lost}, ITC for multilingual real APIs \cite{zhang2026enhancing}, and WildAGTEval for API specification/execution complexity \cite{kim2026beyond}. These works improve coverage and realism, but still mostly report task success rather than concise failure attribution.

% \paragraph{Tool-use and function-calling benchmarks.}
% Early work studies tool acquisition from demonstrations, documentation,
% or self-supervised annotations
% \cite{schick2023toolformer,patil2024gorilla,li2023apibank,tang2023toolalpaca},
% while ToolLLM scales to thousands of real-world RESTful APIs
% \cite{qin2024toolllm}. Recent work emphasizes executable evaluation:
% BFCL covers single-turn, parallel, multi-turn, and abstention cases with
% AST-based matching \cite{patil2025berkeley}; StableToolBench ensures
% reproducibility via virtual API servers \cite{guo2024stabletoolbench};
% FuncBenchGen controls dependency depth through hidden function DAGs
% \cite{maekawa2025towards}; HammerBench targets mobile-assistant scenarios
% \cite{wang2024hammerbench}; and ComplexFuncBench extends to multi-step
% constrained calls under long context \cite{zhong2025complexfuncbench}.
% T-Eval further decomposes tool utilization into six sub-capabilities for
% step-by-step profiling \cite{chen2024teval}. However, these benchmarks
% remain primarily outcome-oriented---reporting whether calls succeed or
% which capability is weak---without decomposing failures into mutually
% exclusive error categories at the query level, constructing adversarial
% boundary negatives, or calibrating difficulty through observed routing
% behavior. PluginEval addresses these gaps by combining execution-grounded
% annotation, three types of boundary negatives, behavioral difficulty
% calibration, and gold-anchored error attribution.

\paragraph{Agentic task benchmarks and automatic evaluation.}
Agent benchmarks evaluate interactive and stateful environments. SWE-bench uses real GitHub fixes \cite{jimenez2024swe}; AgentBench spans operating systems, databases, knowledge graphs, games, and web tasks \cite{liu2024agentbench}; AppWorld and ToolSandbox provide executable application and API environments \cite{trivedi2024appworld,lu2025toolsandbox}. $\tau$-bench and $\tau^2$-Bench evaluate service interactions under explicit policies \cite{yao2024tau,barres2025tau}, while EgoBench extends evaluation to egocentric multimodal tool use \cite{liu2026egobench}. LLM-as-a-judge methods, including MT-Bench, Chatbot Arena, and G-Eval, enable scalable semantic evaluation \cite{zheng2023judging,liu2023g}. Together, these benchmarks improve realism and reduce cost, but mainly assess final outcomes and offer limited diagnosis of routing failures. In contrast, we focus on constructing difficult samples and attributing errors against gold annotations.

% ~0.5 page, minimal.

% 2.1 FC evaluation benchmarks: BFCL (AST/executable, no parameter-level
% attribution); tau/tau^2 (real interaction, only two coarse attribution
% classes); FuncBenchGen (failure types on synthetic variable graphs).
% Point out each one's attribution / reliability gap in one or two sentences.
% TODO.

% 2.2 MCP and agentic evaluation: MCP-Bench / MCP-Atlas (real tools,
% task-completion / claim-coverage oriented, not call-level attribution) --
% one sentence of demarcation. TODO.

% 2.3 Others in brief: capability dimensions (T-Eval); complexity / nesting
% (Nexus / ComplexFuncBench); synthetic data (ToolBench / APIGen / ToolACE),
% one sentence each. TODO.

% (The feature-comparison table may go here or at the start of Section 3.)

\begin{figure*}[t]
\centering
\includegraphics[width=0.98\textwidth]{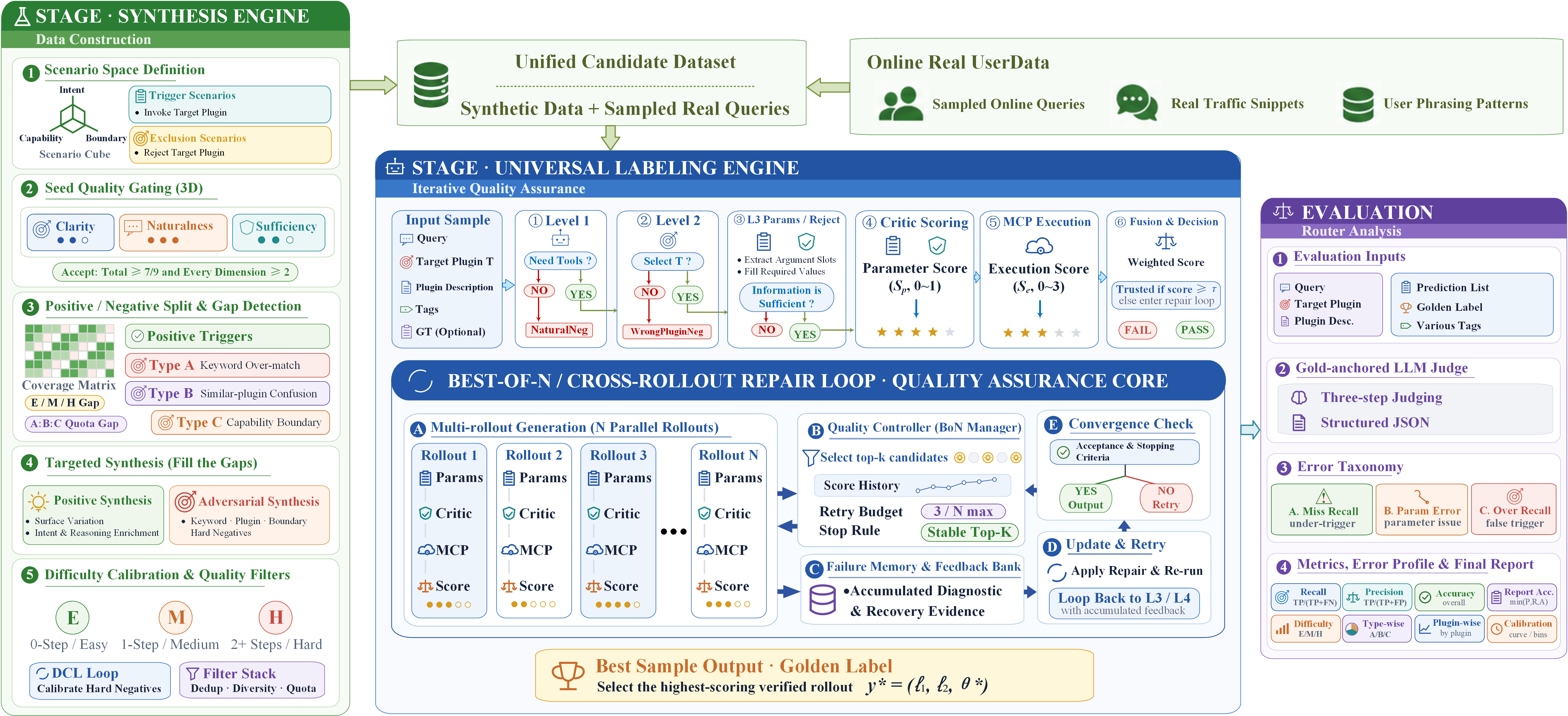}
\caption{Overview of PluginEval. PCCF combines data synthesis guided by scenarios, hierarchical annotation grounded in execution, and iterative quality assurance to construct reliable benchmark instances. The evaluation framework uses gold annotations to measure model performance and attribute failures to missed calls, parameter errors, and spurious calls.}
\label{fig:framework}
\end{figure*}

\section{PluginEval Closed-Loop Construction Framework}
\subsection{Framework Overview and Task Formulation}
\label{sec:framework_overview}

We introduce the PluginEval Closed-Loop Construction Framework (PCCF), an iterative framework for constructing tool routing benchmarks. PCCF comprises two connected stages. Stage I produces annotations verified through execution. Stage II identifies coverage gaps, synthesizes targeted instances, calibrates their difficulty, and returns them to Stage I for verification.

Given a query $q$ and a target plugin $t$, let $d_t$, $\mathcal{S}_t$, and $\Theta_t$ denote the plugin description, argument schema, and space of valid arguments under the schema, respectively. We formulate tool routing as a hierarchy of three decisions:$f_1(q)\in\{0,1\}$,$f_2(q,t,d_t)\in\{0,1\}$,$f_3(q,t,d_{t},\mathcal{S}_t)\in\Theta_t\cup\{\bot\}$.
First, $f_1$ determines whether the query requires a tool. If it does, $f_2$ determines whether the target plugin supports the requested capability. If both decisions are positive, $f_3$ constructs a valid invocation from the information provided in the query. It returns $\bot$ if the query lacks sufficient information to instantiate all required arguments.

Since relevance depends on the target plugin, we evaluate $f_2$ independently for each pair $(q,t)$. Each pair is annotated with $y=(\ell_1,\ell_2,\theta^*)$, where $\ell_1$ indicates whether a tool is necessary, $\ell_2$ indicates whether the target plugin is relevant, and $\theta^* \in \Theta_t \cup \{\bot\}$ denotes the gold invocation. The label space is
$\mathcal{Y}=\{(0,0,\bot),(1,0,\bot),(1,1,\bot)\}
\cup(\{(1,1)\}\times\Theta_t)$,
corresponding to unnecessary tool use, an irrelevant target plugin, a relevant but relevant but underspecified request, and a valid invocation, respectively. By convention, $\ell_2=0$ when $\ell_1=0$, as plugin relevance is not assessed.

\subsection{Stage I: Execution Grounded Ensemble Annotation}
\label{sec:stage1}

% Stage I applies the same annotation pipeline to real queries and synthetic queries returned by Stage II. Real queries preserve natural usage patterns, whereas synthetic queries expand long tail scenarios and capability boundaries. For each query that passes both routing decisions, $M$ generators independently produce complete invocation proposals\citep{wang2022self}. At refinement round $k$, generator $m$ outputs
% \begin{equation}
% z_m^{(k)}
% \in
% \widetilde{\Theta}_t\cup\{\bot\},
% \label{eq:proposal_space}
% \end{equation}
% where $\widetilde{\Theta}_t$ is the unconstrained space of candidate argument assignments and $\bot$ denotes abstention. PCCF evaluates each proposal atomically and never combines argument fields across generators.

Stage I applies a shared annotation pipeline to both real queries and synthetic queries generated by Stage II. Real queries reflect natural usage patterns, while synthetic queries extend coverage to long tail scenarios and capability boundaries. For each query with positive decisions at the first two routing levels, $M$ generators independently propose complete invocations \citep{wang2022self}. At refinement round $k$, generator $m$ produces
\begin{equation}
z_m^{(k)} \in \widetilde{\Theta}_t \cup \{\bot\},
\label{eq:proposal_space}
\end{equation}
where $\widetilde{\Theta}_t$ denotes the unconstrained space of candidate argument assignments, and $\bot$ denotes abstention. PCCF evaluates each proposal in its entirety and does not combine argument fields from different generators.

The verification operator processes proposals according to their type. For an executable proposal, it first applies deterministic schema validation. A schema valid proposal is scored by an argument evaluator, which checks whether its arguments are supported by the query and consistent with the plugin semantics. The proposal is then executed through MCP. If MCP returns a valid response, a separate result evaluator checks whether the response satisfies the user request. Its final score is
\begin{equation}
s_m^{(k)}
=
\lambda_{\mathrm{arg}}
S_{\mathrm{arg}}
\left(
z_m^{(k)}\mid q,t
\right)
+
\lambda_{\mathrm{out}}
S_{\mathrm{out}}
\left(
o_m^{(k)}\mid q,t
\right)
\end{equation}
Both $S_{\mathrm{arg}}$ and $S_{\mathrm{out}}$ lie in $[0,1]$. We require $\lambda_{\mathrm{arg}},\lambda_{\mathrm{out}}\geq 0$ and $\lambda_{\mathrm{arg}}+\lambda_{\mathrm{out}}=1$, ensuring that $s_m^{(k)}\in[0,1]$. Here, $o_m^{(k)}$ denotes the observed MCP response. Proposals with schema violations, execution errors, timeouts, or empty responses are ineligible, but their traces are retained for refinement. Abstention proposals are not executed. Instead, the argument evaluator determines whether missing information or unsupported capability precludes a valid invocation. A justified abstention remains eligible and receives a calibrated score. Both proposal types use the same score range and acceptance threshold $\tau$. Separating generation from evaluation reduces direct self evaluation bias, although LLM evaluators may retain systematic biases \citep{liu2023g,zheng2023judging,ye2025justice}. This design builds on verification pipelines combining structural checks, model based evaluation, and execution feedback \citep{liu2024apigen,liu2025toolace}.

In each round, Stage~I selects the eligible proposal with the highest score. If the score reaches $\tau$, it becomes the gold invocation target. Otherwise, proposals and evidence are appended to the refinement history and supplied to each generator in the next round. Unlike standard self refinement \citep{madaan2023self}, PCCF uses independent evaluator assessments, deterministic validation errors, and observed execution outcomes as repair signals. Refinement stops upon acceptance or after $R_{\max}$ rounds. 

\subsection{Stage II: Scenario Grounded Synthesis and Behavioral Calibration}
\label{sec:stage2}

Stage II derives a scenario catalog from each plugin description. Trigger scenarios represent supported requests that should invoke the plugin. Exclusion scenarios include requests that require no tool, require another plugin, or exceed the target plugin's capabilities. Each example verified in Stage I is assigned one primary scenario for coverage accounting, with other matches retained as auxiliary tags.

Coverage units are further stratified by difficulty. For positive examples, difficulty is determined by the minimum inference depth required for correct routing. Zero, one, and at least two steps correspond to Easy, Medium, and Hard, respectively \citep{wei2022chain}. For negative examples, the construction target $d_{\mathrm{syn}}$ depends on the negative type: domain distance for Type A, surface similarity for Type B, and capability boundary evidence for Type C. At round $r$, the deficit for unit $u$ is
\begin{equation}
\Delta_r(u)=\max\{0,b(u)-n_r(u)\},
\label{eq:coverage_deficit}
\end{equation}
where $n_r(u)$ and $b(u)$ denote its verified count and target budget, respectively. Stage II synthesizes examples only for units with $\Delta_r(u)>0$.

Generated candidates are filtered and returned to Stage I for verification. Verified positives retain their inference difficulty. Verified negatives are assessed by a frozen reference router over $K$ controlled runs \citep{yao2024tau}:
\begin{equation}
p_K(x)
=
\frac{1}{K}
\sum_{j=1}^{K}
\mathbf{1}
_{\left\{
\widehat{y}^{(j)}(x)=y(x)
\right\}}.
\label{eq:dcl}
\end{equation}
The behavioral difficulty $d_{\mathrm{beh}}(x)$ is Hard if $p_K(x)=0$, Medium if $0<p_K(x)<1$, and Easy if $p_K(x)=1$ \citep{polo2024tinybenchmarks}. If the calibrated difficulty is lower than the target, recognition evidence is used to strengthen the negative before further verification and calibration. Only verified examples contribute to the coverage unit corresponding to their observed difficulty.

\subsection{Closed Loop Integration and Extensibility}
\label{sec:closed_loop}

PCCF couples the two stages through verified examples and structured feedback. Stage I returns annotations, execution evidence, and failure traces. Stage II uses these outputs to update coverage, synthesize missing scenarios, and strengthen weak negatives before returning new queries to Stage I. The loop terminates when all coverage budgets are met, verified coverage remains below an improvement threshold for $L$ consecutive rounds, or the synthesis budget is exhausted.

This decomposition separates two questions often conflated in benchmark construction: whether an example is trustworthy and whether it fills an identified coverage gap. Stage I establishes trustworthiness through verification grounded in execution. Stage II determines its necessity through scenario profiling and coverage analysis. Each accepted example is thus directly linked to a coverage gap, a verification record, and a calibrated difficulty label. Given a plugin description, schema, executable interface, and capability boundaries, PCCF requires no specific router architecture. It transforms single pass generation and filtering into an iterative process of scenario discovery, verification through execution, and refinement based on router behavior.

\section{Gold-Anchored Evaluation Framework}

%\subsection{Gold-Anchored Formulation}
\subsection{Semantic Consistency Criterion}

Judging the correctness of tool-call arguments requires determining whether the arguments of matched predicted calls are semantically consistent with their gold counterparts and the intent expressed in the user query. Exact string matching is inadequate for this purpose, as semantically equivalent arguments may differ in surface form. To improve judgment consistency and reproducibility, we introduce a \emph{gold-anchored} LLM judge that takes the predicted call list \(\mathcal{P}\), the human-verified gold call list \(\mathcal{G}\), and the user query as input, while treating \(\mathcal{G}\) as the sole evaluation anchor. The judge uses the query only to interpret semantic equivalence and is explicitly prohibited from extending or overriding the gold specification with additional requirements. Consequently, a predicted call is not considered correct merely because it appears plausible; its arguments must be semantically consistent with the corresponding gold arguments under the structured evaluation procedure described below.

\subsection{Hierarchical Tool-Call Verification}
\label{sec:judge-procedure}

Our gold-anchored LLM judge verifies tool calls hierarchically, proceeding from tool alignment to argument filtering and semantic verification. This ordering ensures that only comparable calls enter argument evaluation and prevents tool selection errors from being conflated with parameter errors.

\paragraph{Stage 1: Tool alignment.}
Let $\mathcal{T}_{P}$ and $\mathcal{T}_{G}$ denote the sets of tool names in the predicted and gold call lists, respectively. The judge matches tools by exact name and retains their intersection, $\mathcal{I}=\mathcal{T}_{P}\cap\mathcal{T}_{G}$, for argument verification. Tools appearing only in the prediction, $\mathcal{T}_{P}\setminus\mathcal{T}_{G}$, or only in the gold calls, $\mathcal{T}_{G}\setminus\mathcal{T}_{P}$, are recorded as tool selection mismatches.

\paragraph{Stage 2: Argument filtering.}
For each tool in $\mathcal{I}$, the judge checks whether the predicted and gold calls contain valid, nonempty argument sets. If either side is empty or missing, the pair is excluded from semantic verification and recorded as a skipped case. This step restricts argument evaluation to calls for which a meaningful comparison can be made.

\paragraph{Stage 3: Semantic verification.}
For each remaining pair, the LLM judge determines whether the predicted arguments are semantically equivalent to the gold arguments under the user query. It returns a binary verdict $r\in\{0,1\}$ together with a brief justification. The comparison follows semantic intent rather than surface form, allowing aliases, formatting variations, equivalent numeric or type representations, and list relations that preserve the requested constraint. Temporal scope is evaluated strictly; for example, an interval with distinct endpoints is not equivalent to a single gold date.

The procedure preserves the source of each failure. Unmatched tools indicate tool selection mismatches, skipped pairs identify calls without comparable arguments, and binary verdicts capture semantic inconsistencies among aligned calls. These outputs support fine-grained error attribution rather than collapsing structurally different failures into a single incorrect prediction.

\subsection{Query-Level Evaluation and Error Attribution}
\label{sec:query-evaluation}

We evaluate function calling at the query level because a request may require multiple coordinated tool calls. A positive query is considered successful only when every required tool is recalled and all arguments are correct. Partial completion is therefore treated as unsuccessful, preventing incomplete multi-tool predictions from inflating performance.

We construct a query-level confusion matrix by sample type. A true positive ($TP$) is a positive query completed correctly, whereas a false negative ($FN$) is a positive query containing a missing tool, an incorrect argument, or a judge failure. For a targeted negative query, invoking the designated target tool produces a false positive ($FP$); otherwise, it is a true negative ($TN$). For a global negative query, invoking any tool produces an $FP$, while producing no tool call yields a $TN$. Cases marked as both\_empty are excluded.

We report query-level Recall,
$\mathrm{Recall}=TP/(TP+FN)$,
and Accuracy,
$\mathrm{Accuracy}=(TP+TN)/(TP+FN+TN+FP)$.
Under our strict query-level criterion, Recall measures end-to-end success on positive queries rather than tool-name recall alone, while Accuracy measures correctness across both positive and negative queries.

Each failed query is further assigned to a mutually exclusive error category. \emph{Miss recall} denotes missing required tools, \emph{parameter error} denotes semantically incorrect arguments for an aligned tool, and \emph{over recall} denotes an erroneous tool activation on a negative query. Parameter errors are further classified as temporal errors, missing arguments, or incorrect values according to the judge justification, while judge failures are recorded separately. Accordingly, false negatives consist of miss recall, parameter errors, and judge failures, whereas false positives correspond to over recall.

\begin{table*}[t]
\centering
\normalfont
\fontsize{10}{12}\selectfont
\setlength{\tabcolsep}{3pt}
\renewcommand{\arraystretch}{0.96}
\setlength{\aboverulesep}{1pt}
\setlength{\belowrulesep}{1pt}

\begin{tabular}{@{}lcccccccc@{}}
\toprule
& \multicolumn{2}{c}{\textbf{Performance}}
& \multicolumn{5}{c}{\textbf{Positive Query Errors}}
& \multicolumn{1}{c}{\textbf{Negative Query Errors}} \\
\cmidrule(lr){2-3}
\cmidrule(lr){4-8}
\cmidrule(lr){9-9}

\textbf{Model}
& \textbf{Recall} $\uparrow$
& \textbf{Acc.} $\uparrow$
& \textbf{Miss} $\downarrow$
& \textbf{Param.} $\downarrow$
& \textbf{Temp.} $\downarrow$
& \textbf{Missing} $\downarrow$
& \textbf{Value} $\downarrow$
& \textbf{Over} $\downarrow$ \\
\midrule

\addlinespace[0pt]
\multicolumn{9}{@{}l}{
Original distribution
(Easy/Medium/Hard $=0.40/0.40/0.20$)
} \\
\addlinespace[0pt]

GPT-5.4
& 0.4920 & 0.5213 & 0.3372 & 0.1708
& 0.0796 & 0.0540 & 0.0372 & 0.3320 \\

Claude Opus 4.6
& \textbf{0.6760} & \textbf{0.6817}
& 0.1395 & 0.1845 & 0.0936 & 0.0370 & 0.0539 & 0.2910 \\

Gemini 3.1 Pro
& \underline{0.6496} & \underline{0.6778}
& 0.1766 & 0.1738 & 0.0897 & 0.0434 & 0.0406 & 0.1855 \\

DeepSeek V4 Pro
& 0.6294 & 0.6433 & 0.1829 & 0.1877
& 0.0949 & 0.0466 & 0.0462 & 0.2891 \\

Qwen3-235B-A22B
& 0.5482 & 0.5840 & 0.2428 & 0.2090
& 0.1041 & 0.0583 & 0.0466 & 0.2422 \\

\midrule

\addlinespace[0pt]
\multicolumn{9}{@{}l}{
Easy-weighted distribution
(Easy/Medium/Hard $=0.60/0.30/0.10$)
} \\
\addlinespace[0pt]

GPT-5.4
& 0.7035 & 0.7462 & 0.2157 & 0.0808
& 0.0446 & 0.0217 & 0.0145 & 0.0419 \\

Claude Opus 4.6
& \textbf{0.8233} & \textbf{0.8335}
& 0.0833 & 0.0933 & 0.0429 & 0.0244 & 0.0260 & 0.1158 \\

Gemini 3.1 Pro
& \underline{0.7973} & \underline{0.8198}
& 0.1074 & 0.0954 & 0.0445 & 0.0244 & 0.0264 & 0.0692 \\

DeepSeek V4 Pro
& 0.7852 & 0.8024 & 0.1158 & 0.0990
& 0.0409 & 0.0285 & 0.0297 & 0.1118 \\

Qwen3-235B-A22B
& 0.7162 & 0.7413 & 0.1631 & 0.1206
& 0.0497 & 0.0425 & 0.0285 & 0.1337 \\

\bottomrule
\end{tabular}

\caption{Performance and error rates under the original and easy-weighted difficulty distributions. The best and second-best Recall and Accuracy results in each setting are shown in bold and underlined, respectively.}
\label{tab:main-results}
\end{table*}

\begin{table*}[t]
\centering
\small
\setlength{\tabcolsep}{2pt}
\renewcommand{\arraystretch}{0.96}
\setlength{\aboverulesep}{1pt}
\setlength{\belowrulesep}{1pt}
\begin{tabular}{@{}lcccccccc@{}}
\toprule
& \multicolumn{2}{c}{\textbf{PluginEval}}
& \multicolumn{2}{c}{\textbf{BFCL}}
& \multicolumn{2}{c}{\textbf{APIBench}}
& \multicolumn{2}{c}{\textbf{NexusRaven}} \\
\cmidrule(lr){2-3} \cmidrule(lr){4-5} \cmidrule(lr){6-7} \cmidrule(lr){8-9}
\textbf{Model}
& \textbf{Rec.} & \textbf{Acc.}
& \textbf{Rec.} & \textbf{Acc.}
& \textbf{Rec.} & \textbf{Acc.}
& \textbf{Rec.} & \textbf{Acc.} \\
\midrule
DeepSeek-V4-Pro & 0.6294 & 0.6433 & 0.6942 & 0.7370 & 0.4986 & 0.6440 & 0.5278 & 0.6667 \\
GPT-5.4         & 0.4920 & 0.5213 & 0.6000 & 0.6860 & 0.5943 & \underline{0.7090} & 0.7535 & 0.8260 \\
Gemini-3.1-Pro  & \underline{0.6496} & \underline{0.6778} & \underline{0.7029} & \underline{0.7420} & \textbf{0.6643} & \textbf{0.7210} & \textbf{0.8056} & \textbf{0.8627} \\
Claude-Opus-4.6 & \textbf{0.6760} & \textbf{0.6817} & \textbf{0.7304} & \textbf{0.7490} & \underline{0.6514} & 0.7060 & \underline{0.7569} & \underline{0.8284} \\
\bottomrule
\end{tabular}
\caption{Recall and Accuracy on PluginEval and three existing function calling benchmarks. The best and second best results for each benchmark are shown in bold and underlined, respectively.}
\label{tab:main_recall_acc}
\end{table*}

% Requires: \usepackage{booktabs}

% Requires: \usepackage{booktabs}
\begin{table*}[t]
\centering
\small
\setlength{\tabcolsep}{2pt}
\renewcommand{\arraystretch}{0.96}
\setlength{\aboverulesep}{1pt}
\setlength{\belowrulesep}{1pt}

\begin{tabular}{@{}lllcccc@{}}
\toprule
\# & Component & Metric & Full & w/o & $\Delta$ & $p$ \\
\midrule
\multicolumn{7}{l}{\textit{Component ablation}} \\
1 & Content Validation
  & FP$\downarrow$
  & \textbf{0\%}
  & 78\%
  & $+78.0$
  & $<\!0.001$ \\

2 & Calibration ($R{=}3$)
  & Pass$\uparrow$
  & \textbf{39.7\%}
  & 11.1\%
  & $-28.7$
  & $<\!0.001$ \\

3 & Closed loop$^{\ddagger}$
  & Pass$\uparrow$
  & \textbf{64\%}
  & 27\%
  & $-37.0$
  & --- \\

  &
  & Cov$\uparrow$
  & \textbf{3.88$\times$}
  & 1.38$\times$
  & $-64\%$
  & --- \\

  &
  & Diff$\downarrow$
  & \textbf{0.40}
  & 1.16
  & $+0.76$
  & --- \\

4 & BoN ($M{=}3$)
  & Pass$\uparrow$
  & \textbf{17.1\%}
  & 12.0\%
  & $-5.2$
  & $<\!0.001$ \\

\midrule
\multicolumn{7}{l}{\textit{Robustness \& sensitivity}} \\

5 & Plugin Scale ($N{=}1{\to}8$)
  & Pass$\uparrow$
  & \textbf{22.3\%}
  & 17.1\%
  & $-5.2$
  & $<\!0.001$ \\

6 & Hyperparam
  & \multicolumn{3}{c}{
      $R_{\max}$: 28.7\,pp $\gg$ $M$: 5.2\,pp $>$ $K$: $\sim$0\,pp
    }
  & \multicolumn{2}{c}{
      Optimal: $R{=}3,\;M{=}3,\;K{\geq}5$
    } \\
\bottomrule
\end{tabular}

\caption{Ablation and sensitivity analysis of PCCF components.}
\label{tab:ablation}
\end{table*}

\section{Experiments}
% ~1.5 pages.

\subsection{Experimental Setup}
\paragraph{Benchmarks.}
Our primary evaluation uses \textsc{PluginEval}, a Chinese function calling benchmark with $3{,}000$ human verified instances covering $54$ plugins, $63$ tools, and nine domains. It contains $2{,}500$ positive queries requiring correct tool calls and grounded arguments, and $500$ adversarial negatives requiring abstention. Easy, medium, and hard instances follow a $0.4\!:\!0.4\!:\!0.2$ ratio. We additionally compare model performance on BFCL~\citep{patil2025berkeley}, APIBench~\citep{patil2024gorilla}, and NexusRaven~\citep{srinivasan2023nexusraven} using their respective evaluation protocols.

\paragraph{Models.}
We evaluate five frontier large language models: GPT-5.4~\cite{singh2025openai}, Claude 4.6~\cite{anthropic2026claudeopus46}, Gemini 3.1 Pro~\cite{team2023gemini}, DeepSeek-V4-Pro~\cite{xu2026deepseek}, and Qwen3-235B-A22B~\cite{yang2025qwen3}. These models cover major proprietary and open-weight model families, enabling a broad comparison of current function-calling capabilities.

\paragraph{Judge Configuration.}
We use GPT-5.2 as the fixed LLM judge for all experiments, with the same prompt and inference configuration applied to every evaluated model. Failed API calls or invalid JSON outputs are retried up to five times.

\paragraph{Metrics.}
We report Recall and Accuracy at the query-level, together with breakdowns of miss recall, parameter errors, and over recall. Parameter errors are further grouped into temporal errors, missing arguments, and incorrect values.

\begin{figure}[t] \centering \includegraphics[width=0.65\columnwidth]{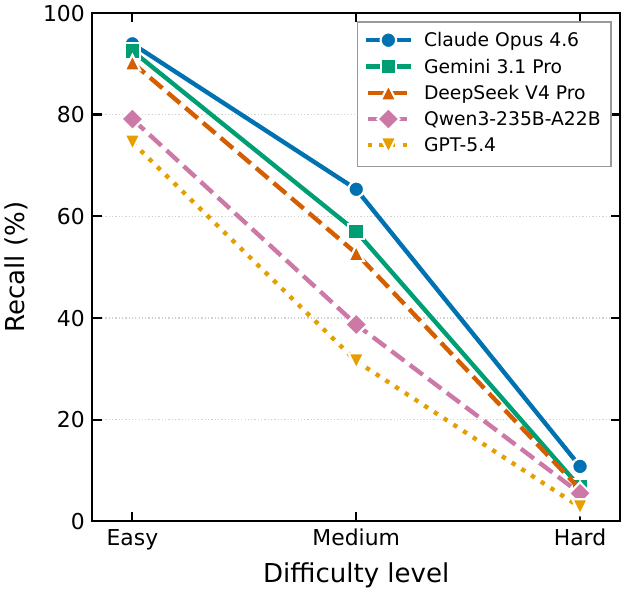} \caption{Recall across difficulty levels. All models decrease monotonically from Easy to Hard, with the largest cross-model gap on Medium instances.}
\label{fig:difficulty} \end{figure}

\subsection{Overall Performance across Difficulty Distributions}
\label{sec:main-results}

Table~\ref{tab:main-results} compares five models under two difficulty distributions, while Figure~\ref{fig:difficulty} reports Recall at each difficulty level. The original distribution contains easy, medium, and hard instances at a ratio of $0.40/0.40/0.20$, whereas the easy-weighted distribution uses $0.60/0.30/0.10$. Recall measures the proportion of positive queries for which all required tools are recalled and all arguments are correct, while Acc. denotes query-level Accuracy over positive and negative queries. Miss, Param., and Over denote miss recall, parameter errors, and over recall, respectively. Temp., Missing, and Value further divide parameter errors into temporal errors, missing arguments, and incorrect values. Positive query errors are normalized by the number of positive queries, whereas Over is normalized by the number of negative queries.

On the original evaluation set, Claude Opus 4.6 achieves the strongest overall performance, with $67.60\%$ Recall and $68.17\%$ Accuracy. Gemini 3.1 Pro ranks second, reaching $64.96\%$ Recall and $67.78\%$ Accuracy, followed by DeepSeek V4 Pro. Figure~\ref{fig:difficulty} further reveals a sharp decline as difficulty increases. Recall ranges from $74.5\%$ to $93.9\%$ on easy instances, falls to $31.5\%$--$65.3\%$ on medium instances, and reaches only $2.8\%$--$10.8\%$ on hard instances. Claude Opus 4.6 performs best at every level, but even its Recall drops to $10.8\%$ on hard queries. The error breakdown also shows that performance is not determined by argument generation alone. GPT-5.4 has the lowest parameter error rate, but its substantially higher miss recall and over recall reduce its overall performance.

Increasing the easy instance proportion from $0.40$ to $0.60$, while reducing the medium and hard proportions, substantially improves every model. Recall rises by $14.73$ to $21.15$ percentage points, and Accuracy rises by $14.20$ to $22.49$ points. Claude Opus 4.6 remains the strongest model, reaching $82.33\%$ Recall and $83.35\%$ Accuracy, followed by Gemini 3.1 Pro and DeepSeek V4 Pro. The model ordering is largely preserved, although GPT-5.4 attains slightly higher Accuracy than Qwen3-235B-A22B despite lower Recall because of its much lower over recall rate. Together with the difficulty-level results, these findings demonstrate that aggregate function calling scores are highly sensitive to the difficulty composition of the evaluation set.

\begin{figure}[t]
\centering
\includegraphics[width=0.8\columnwidth]{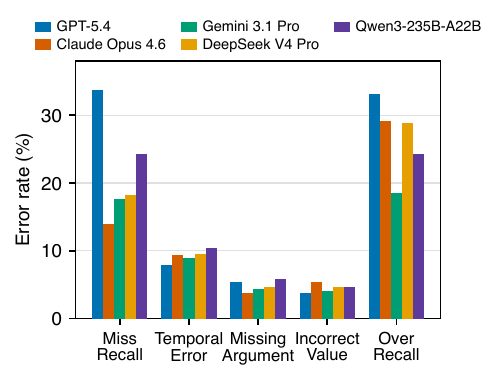}
\caption{Error attribution on the original evaluation set. Positive-query errors and Over Recall are normalized by positive and negative queries, respectively. Lower is better.}
\label{fig:attribution}
\end{figure}

\subsection{Comparison with Existing Function Calling Benchmarks}
\label{sec}

Table~\ref{tab:main_recall_acc} compares model performance on PluginEval with three function calling benchmarks. All four models achieve lower Recall and Accuracy on PluginEval than on BFCL. For example, Claude Opus 4.6 decreases from $73.04\%$ Recall and $74.90\%$ Accuracy on BFCL to $67.60\%$ and $68.17\%$ on PluginEval, respectively. GPT-5.4 shows a larger decrease, from $60.00\%$ Recall and $68.60\%$ Accuracy to $49.20\%$ and $52.13\%$. PluginEval also yields a wider Accuracy range than BFCL and APIBench, revealing more pronounced performance differences among the evaluated models.

Model rankings also vary across benchmarks. Claude Opus 4.6 ranks first on PluginEval and BFCL, whereas Gemini 3.1 Pro performs best on APIBench and NexusRaven. DeepSeek V4 Pro outperforms GPT-5.4 on PluginEval but ranks below it on APIBench and NexusRaven. These ranking shifts indicate that model performance depends strongly on the capabilities and failure modes emphasized by each benchmark. PluginEval complements existing evaluations by placing greater emphasis on adversarial negatives, hierarchical routing decisions, and strict argument grounding.

\subsection{Fine-Grained Error Attribution}
\label{sec:error-attribution}

Figure~\ref{fig:attribution} reveals distinct tool selection and rejection behaviors across models. GPT-5.4 exhibits the highest Miss Recall and Over Recall rates, reaching $33.72\%$ and $33.20\%$, respectively, indicating substantial difficulty in both identifying required tools and avoiding unnecessary calls. Qwen3-235B-A22B shows a similar but less pronounced pattern. Claude Opus 4.6 achieves the lowest Miss Recall rate at $13.95\%$, demonstrating strong coverage of required tools, but its Over Recall remains high at $29.10\%$. In contrast, Gemini 3.1 Pro attains the lowest Over Recall rate of $18.55\%$, which helps explain its competitive Accuracy despite a lower Recall than Claude Opus 4.6.

At the argument level, Temporal Error is the largest parameter error subtype for every model, ranging from $7.96\%$ to $10.41\%$, and consistently exceeds both Missing Argument and Incorrect Value. This identifies temporal grounding as a shared weakness across model families. The results also show that similar aggregate performance can arise from different failure sources: GPT-5.4 and Qwen3 are primarily limited by tool selection, Claude Opus 4.6 combines strong recall with weaker abstention, while Gemini 3.1 Pro exhibits the most balanced error profile.

\subsection{Judge Reliability Analysis}
% 5.4 The linchpin -- keep it prominent.
%   - judge attribution vs human attribution agreement (Cohen's / Fleiss' kappa);
%   - judge ranking vs human ranking correlation (Spearman rho);
%   - stability / self-consistency (repeated runs, swapping judge models);
%   - comparison against AST / naked LLM-judge baselines (more accurate, more stable).
% TODO.
\label{sec:judge-reliability}

We first audit the reference annotations, not model outputs. From the final evaluation set of 3,000 cases, containing 2,500 positives and 500 negatives, we sample 300 cases with seed 20260725 and preserve the original class prior, yielding 250 positives and 50 negatives. Three human reviewers assess plugin necessity on all cases, and target relevance and argument grounding on positives.

\begin{table}[t]
\centering
\small
\setlength{\tabcolsep}{2pt}
\setlength{\aboverulesep}{1pt}
\setlength{\belowrulesep}{1pt}
\begin{tabular}{@{}lccccc@{}}
\toprule
Metric & $N$ & R1 & R2 & R3 & $\kappa_F$ \\
\midrule
Need & 300 & 99.00\% & 100.00\% & 99.33\% & 0.3966 \\
Target & 250 & 98.80\% & 98.80\% & 98.80\% & 1.0000 \\
Args. & 250 & 98.00\% & 96.40\% & 97.60\% & 0.6918 \\
Overall & 300 & 96.67\% & 96.33\% & 96.67\% & 0.6993 \\
\bottomrule
\end{tabular}
\caption{Human audit of reference quality. Entries are agreement rates; $\kappa_F$ is Fleiss' $\kappa$.}
\label{tab:gt-quality}
\end{table}

Table~\ref{tab:gt-quality} reports high reference quality along the three decision levels in our formulation. Need audits whether the target plugin should be invoked over both positive and negative cases; Target checks whether the intended plugin is matched on positives; Args. checks whether grounded arguments are correct on positives. Overall validity is 96.67\%, 96.33\%, and 96.67\% for the three reviewers, respectively. Fleiss' $\kappa$ is highest for Target because all reviewers agree on the same target-selection errors; the lower Need $\kappa$ reflects the rarity of negative labels under high marginal agreement.

We then audit the judge on a separate stratified subset of 300 cases: 100 correct predictions, 50 missed calls, 50 spurious calls, and 100 parameter errors. Reviewers verify whether each judge attribution matches the gold outcome.

\begin{table}[t]
\centering
\small
\setlength{\tabcolsep}{2pt}
\setlength{\aboverulesep}{1pt}
\setlength{\belowrulesep}{1pt}
\begin{tabular}{@{}lcccc@{}}
\toprule
Metric & $N$ & R1 & R2 & R3 \\
\midrule
Correct & 100 & 95.00\% & 99.00\% & 95.00\% \\
Miss & 50 & 100.00\% & 100.00\% & 100.00\% \\
Over & 50 & 100.00\% & 100.00\% & 100.00\% \\
Param. & 100 & 93.00\% & 92.00\% & 88.00\% \\
\midrule
Agreement ($P_o$) & 300 & 96.00\% & 97.00\% & 94.33\% \\
Cohen's $\kappa$ & 300 & 0.9451 & 0.9585 & 0.9221 \\
\bottomrule
\end{tabular}
\caption{Human audit of judge reliability. Miss, Over, and Param. denote missed-call, spurious-call, and parameter-error categories.}
\label{tab:judge-reliability}
\end{table}

Table~\ref{tab:judge-reliability} shows strong alignment between human review and judge attribution. The judge reaches 96.00\%, 97.00\%, and 94.33\% observed agreement with the three reviewers, while reviewer-specific Cohen's $\kappa$ values are 0.9451, 0.9585, and 0.9221 after correcting for chance agreement. Separately, the three human reviewers show high inter-reviewer agreement, with Fleiss' $\kappa=0.9451$. All reviewers confirm all missed call and spurious call attributions; the remaining disagreements are concentrated in parameter errors, where confirmation is 93.00\%, 92.00\%, and 88.00\%.

\subsection{Ablation Study}
Table~\ref{tab:ablation} demonstrates that the four PCCF components address complementary aspects of benchmark construction. Removing content validation raises the false positive rate from $0\%$ to $78\%$, confirming that schema valid and executable calls may still return irrelevant results. Disabling iterative calibration reduces the cumulative pass rate from $39.7\%$ to $11.1\%$, showing that feedback based refinement recovers many initially invalid but correctable samples. For the complex plugin setting, replacing closed loop synthesis with single shot generation lowers the pass rate from $64\%$ to $27\%$, reduces scenario coverage from $3.88\times$ to $1.38\times$, and increases difficulty deviation from $0.40$ to $1.16$. Reducing BoN sampling from $M=3$ to $M=1$ further decreases the pass rate from $17.1\%$ to $12.0\%$. The effects of content validation, iterative calibration, and BoN sampling are statistically significant, while the benefits of content validation and BoN remain consistent from one to eight plugins. Sensitivity analysis further identifies $R=3$, $M=3$, and $K\geq5$ as suitable settings. Overall, these results show that PCCF jointly improves data quality, effective yield, scenario coverage, and difficulty control.

\section{Conclusion}
% We presented PluginEval, a Chinese benchmark for fine grained evaluation of function calling. Its PCCF framework combines execution grounded annotation, scenario guided synthesis, and behavioral difficulty calibration, while the gold anchored judge attributes failures to missed calls, spurious calls, and parameter errors. Experiments on five frontier LLMs reveal strong sensitivity to difficulty and distinct error profiles hidden by aggregate scores. Human audits and ablations further confirm the reliability of the benchmark, judge, and construction framework.
We introduced PluginEval, a Chinese benchmark for diagnostic function calling evaluation. PCCF combines execution evidence for annotation, scenario specifications for synthesis, and model behavior for difficulty calibration. Its judge uses gold labels to attribute missed calls, spurious calls, and parameter errors. Across five LLMs, results show that test composition affects measured performance, while aggregate scores obscure routing failures. Human audits and ablations support the design. These findings motivate structured error attribution beyond aggregate success.

% References and End of Paper
% These lines must be placed at the end of your paper.
\bibliography{aaai2027}

% Check whether the conference requires a reproducibility checklist to be included in the paper.
% If so, you can uncomment the following line and ajust the path to include it.
% \input{ReproducibilityChecklist.tex}

\end{document}